\documentclass[11pt,a4paper]{article}
 
\usepackage[T1]{fontenc}
\usepackage[utf8]{inputenc}
\usepackage[english]{babel}
 
\usepackage[a4paper, top=2.5cm, bottom=2.5cm, left=2.5cm, right=2.5cm]{geometry}
 
\usepackage{amsmath,amssymb}
\usepackage{graphicx}
\usepackage{array}
\usepackage{caption}
\usepackage{times}
 
\usepackage[round,authoryear]{natbib}
\setcitestyle{aysep={,}}
\usepackage{hyperref}
\hypersetup{
  colorlinks=false,
  citebordercolor={1 0 0},
  linkbordercolor={1 0 0},
  urlbordercolor={1 0 0},
  pdfborder={0 0 1}
}
\usepackage{xstring}
\newcommand{\doi}[1]{%
  doi:~%
  \IfBeginWith{#1}{10.5555/}%
    {\href{https://dl.acm.org/doi/#1}{#1}}%
    {\href{https://doi.org/#1}{#1}}%
}
 
\usepackage{sectsty}
\allsectionsfont{\sffamily\bfseries}
 
\usepackage{titling}
\pretitle{\begin{flushleft}\Large\bfseries}
\posttitle{\end{flushleft}\vskip 0.5em}
\preauthor{\begin{flushleft}\normalsize}
\postauthor{\end{flushleft}}
\predate{}\postdate{}
\date{}

\begin{document}

\setlength{\droptitle}{-2.5em}
\title{DINOv3-MIL: Per-Kidney Multi-Label Tumour and Cyst Detection from Foundation-Model Patch Tokens on KiTS23}

\author{}
\maketitle
\vspace{-3.5em}

{\raggedright
Vishalakshi M\textsuperscript{1}, Sahil Sharma\textsuperscript{2*}, Pramod Kumar P\textsuperscript{1}

\smallskip
\textsuperscript{1}Department of Computer Science and Artificial Intelligence, SR University, Warangal, India

\textsuperscript{2}School of Computing, Ulster University, Belfast, United Kingdom

\smallskip
\textbf{* Correspondence:} \\
Corresponding Author \\
s.sharma@ulster.ac.uk
\par}

\vspace{0.6em}

\section*{Abstract}

Foundation vision models trained on natural images transfer to medical tasks without domain pre-training, but volumetric classification requires aggregating tens of thousands of patch tokens per study, and the aggregator constrains how the resulting model can be interpreted. We compare three aggregators on identical frozen DINOv3 ViT-H/16+ features for renal tumour/cyst detection on KiTS23 (966 kidneys; n=97 test): a CLS-token linear probe, gated attention multiple instance learning (MIL) over 55,296 patch tokens, and a prototype head following ProtoViT. Attention MIL achieves the highest AUROC for tumour (0.74, 95\% CI 0.64--0.83) and cyst (0.80, 0.70--0.88), with attention enriched 7.5--9.8$\times$ over chance within annotated lesions. The prototype head does not transfer to cyst detection (AUROC 0.51), exposing an interpretability--performance trade-off at this token scale.

\section{Introduction}

Renal CT is routine in oncological work-up, and deployed classifiers are typically explained by post-hoc saliency methods whose faithfulness has been challenged \citep{adebayo2018}. For foundation-model patch features on volumetric CT, the aggregator choice including global pooling, attention, or prototype matching determines whether the resulting classifier is interpretable, whether it can detect small lesions, and whether the two goals trade off. Applying a ViT to 2.5D axial slices of a kidney crop produces $\sim$55K patch tokens per kidney; we ask which aggregator delivers on per-kidney multi-label tumour and cyst detection.

Self-supervised vision foundation models now produce dense patch features that transfer to medical imaging without domain pre-training \citep[DINOv3;][]{simeoni2025}. Attention-based MIL \citep{ilse2018} aggregates patch features for whole-slide and volumetric classification with post-hoc interpretability via attention weights. Prototype-based networks \citep{chen2019, ma2024} offer intrinsic case-based interpretability \citep{rudin2019}. Classification work on KiTS23 has been confined to mass-level binary tumour-vs-cyst given a known lesion \citep{ortlieb2026}; the per-kidney multi-label setting addressed here has not been studied.

This work makes two contributions. (i) A per-kidney multi-label tumour/cyst task on KiTS23 including healthy kidneys. (ii) A matched comparison of three aggregators on frozen DINOv3 patch features: linear probe, attention MIL, ProtoViT-style head.

\section{Methods}

\begin{figure}[htbp]
\centering
\includegraphics[width=\textwidth]{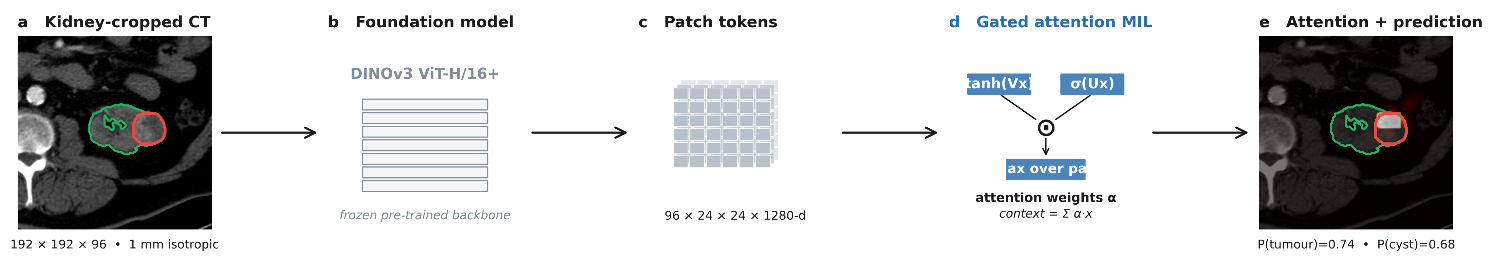}
\caption{Pipeline. (a) Kidney-cropped CT (192$\times$192$\times$96 voxels, 1 mm isotropic) with kidney (green) and tumour (red) annotations. (b) Frozen DINOv3 ViT-H/16+ encodes 2.5D axial triplets. (c) Cached patch tokens of size 96$\times$24$\times$24$\times$1280-d. (d) Gated attention MIL: tanh and sigmoid projections produce attention weights via softmax over 55,296 tokens. (e) Attention-pooled context drives prediction.}
\label{fig:pipeline}
\end{figure}

From 489 KiTS23 cases \citep{heller2023}, kidneys were extracted by connected components, resampled to 1 mm isotropic, intensity-windowed to HU [$-100$, 400], and normalised, yielding 966 crops of 192$\times$192$\times$96 voxels with multi-label (tumour, cyst) at prevalence 0.59 and 0.37. An 80/10/10 case-level split (seed 23) gave 771/98/97 kidneys. Each kidney's 96 axial slices were converted to 2.5D triplets, resized to 384$\times$384, and passed through frozen DINOv3 ViT-H/16+; we cache 576 patch tokens per slice (bf16) (Figure \ref{fig:pipeline}).

Three heads share the cache. The linear probe mean-pools the 96 CLS tokens (lr 1e-3, batch 32, 100 epochs). Gated attention MIL \citep{ilse2018} attention-pools 96$\times$576 = 55,296 tokens via tanh-sigmoid streams (hidden 256, dropout 0.2; lr 5e-4, batch 4, 30 epochs with cosine annealing). The ProtoViT head \citep{ma2024} uses 10 prototypes (5/class) with cosine-similarity max-pooling, a class-constrained classifier, and clustering/separation losses weighted 0.8/0.08 \citep{chen2019}, trained in three stages (5+15+5 epochs). All heads use AdamW (weight decay 1e-4) and BCEWithLogitsLoss with positive-class weighting; seed 23; thresholds tuned by Youden's J; 1000-resample bootstrap CIs.

\section{Results}

Table~\ref{tab:results} reports test performance. Gated attention MIL achieves the highest AUROC for both labels (tumour 0.74, cyst 0.80), with 95\% confidence intervals well above chance. The linear probe approaches MIL on tumour (0.67) but does not transfer to cyst (0.55, CI lower bound 0.42); mean-pooling discards the small-volume signal that spatial attention recovers. The ProtoViT head matches the linear probe on tumour (0.71) but did not exceed chance on cyst (0.51, CI crossing 0.5), despite operating on identical features.

\begin{table}[htbp]
\centering
\caption{Test-set AUROC with 95\% bootstrap CIs (n=97 kidneys); F1 at validation-tuned thresholds.}
\label{tab:results}
\small
\begin{tabular}{@{}lcccc@{}}
\hline
\textbf{Method} & \textbf{Tumour AUROC} & \textbf{Cyst AUROC} & \textbf{Tumour F1} & \textbf{Cyst F1} \\
\hline
Linear probe (CLS) & 0.67 [0.56, 0.77] & 0.55 [0.42, 0.66] & 0.73 & 0.26 \\
ProtoViT head & 0.71 [0.61, 0.81] & 0.51 [0.39, 0.62] & 0.54 & 0.52 \\
Gated attention MIL & 0.74 [0.64, 0.83] & 0.80 [0.70, 0.88] & 0.70 & 0.66 \\
\hline
\end{tabular}
\end{table}

\begin{figure}[htbp]
\centering
\includegraphics[width=\textwidth]{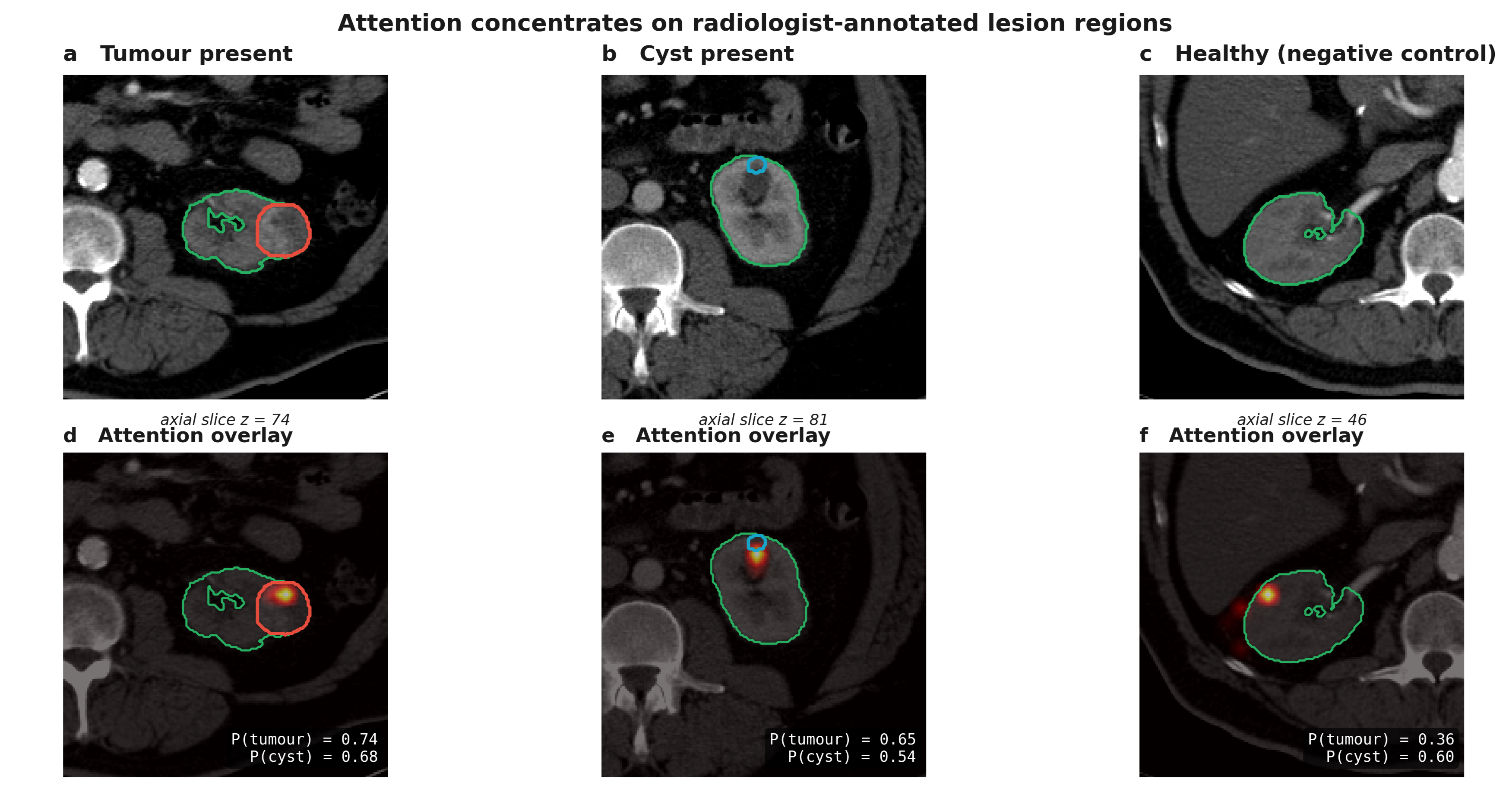}
\caption{Attention overlays on three test kidneys. Top row (a--c): CT slice with annotations. Bottom row (d--f): attention heatmap overlay. Tumour case (z=74): attention concentrates on tumour (red). Cyst case (z=81, within GT span 81--95): attention concentrates near cyst boundary (cyan). Healthy case (z=46): attention diffuses across kidney slices. Quantitative attention--lesion overlap is computed over the full 3D volume.}
\label{fig:attention}
\end{figure}

Figure~\ref{fig:attention} shows attention overlays on three example kidneys. For the tumour case, attention concentrates inside the radiologist-annotated tumour contour (93.3\% of total attention mass inside lesion cells versus 12.4\% expected by chance, 7.5$\times$ enrichment). For the cyst case, 12.4\% of attention lies inside cyst cells against 1.3\% expected by chance (9.8$\times$ enrichment). For the healthy kidney, attention is distributed across slices with no peak exceeding 8\% of total mass with no false concentration on the kidney.

\section{Discussion}

On identical frozen DINOv3 features, gated attention MIL outperforms both mean-pooling and ProtoViT-style prototype learning on per-kidney tumour and cyst detection. The largest gap is on cysts, where attention MIL gains 0.25 AUROC over mean-pooling and 0.29 over the prototype head; we attribute this to spatial reasoning over a large patch budget (55,296 tokens per kidney), which neither global pooling nor max-pool-then-prototype efficiently exploits for small-volume lesions. We note that attention maps localise to radiologist-annotated regions but should not be read as causal explanations of the classifier's decision \citep{adebayo2018}. Limitations include the small held-out set (n=97), a single split, validation-set threshold tuning, and early stopping of both trainable heads. Whether intrinsic prototype interpretability can be recovered at this token scale via relaxed classifier constraints or alternative aggregation remains open.

\renewcommand{\bibsection}{\section{References}}
\bibliographystyle{Frontiers-Harvard}
\bibliography{references}

\section{Conflict of Interest}
The authors declare that the research was conducted in the absence of any commercial or financial relationships that could be construed as a potential conflict of interest.

\section{Author Contributions}
VM conceived the study, implemented the models, conducted the experiments, analysed the results, and wrote the manuscript. SS supervised the implementation, contributed to the narrative, and edited the final manuscript. PKP supervised the overall work. All authors reviewed and approved the final manuscript.

\section{Funding}
This research did not receive any specific grant from funding agencies in the public, commercial, or not-for-profit sectors.

\end{document}